%% file: eccv2016submissionArxiv.tex
\documentclass[runningheads]{llncs}
\usepackage{graphicx}
\usepackage{amsmath,amssymb} %
\usepackage{color}
\usepackage[width=122mm,left=12mm,paperwidth=146mm,height=193mm,top=12mm,paperheight=217mm]{geometry}
\usepackage{xspace}
\usepackage[hidelinks]{hyperref}

\begin{document}
\input{defines}
\input{macros}

\pagestyle{headings}
\mainmatter

\title{Generating Visual Explanations} 

\titlerunning{Generating Visual Explanations}

\authorrunning{L. A. Hendricks, Z. Akata, M. Rohrbach, J. Donahue, B. Schiele, T. Darrell}

 \newcommand{\authSpace}{\ \ \ \ }
\author{%
Lisa Anne Hendricks$^{1}$ \authSpace Zeynep Akata$^{2}$ \authSpace Marcus Rohrbach$^{1,3}$ \\   
Jeff Donahue$^{1}$ \authSpace Bernt Schiele$^{2}$ \authSpace Trevor Darrell$^{1}$\\
}

\institute{
 $^{1}$UC Berkeley EECS,  CA, United States\\
  $^{2}$Max Planck Institute for Informatics, Saarbr{\"u}cken, Germany\\
 $^{3}$ICSI, Berkeley, CA, United States\\
 }

\maketitle

\begin{abstract}
Clearly explaining a rationale for a classification decision to an end-user %
can be as important %
as the decision itself. 
Existing approaches for deep visual recognition are generally opaque and do not output any justification text; contemporary vision-language models can describe image content but fail to take into account class-discriminative image aspects which justify visual predictions.
We  propose a new model that focuses on the discriminating properties of the visible object, jointly predicts a class label, and explains why the predicted label is appropriate for the image. 
We propose a novel loss function based on sampling and reinforcement learning that learns to generate sentences that realize a global sentence property, such as class specificity.  
Our results on a fine-grained bird species classification dataset show that our model is able to generate explanations which 
are not only consistent with an image but also more discriminative than  descriptions produced by existing captioning methods.
\end{abstract}

\section{Introduction}

Explaining why the output of a visual system is compatible with visual evidence is a key component for understanding and interacting with AI systems~\cite{biran2014justification}.
Deep classification methods have had tremendous success in visual recognition \cite{krizhevsky2012imagenet,gao2015compact,donahue2013decaf}, but their predictions can be unsatisfactory if the model cannot provide a consistent justification %
of why it made a certain prediction. 
In contrast, systems which can justify why a prediction is consistent with visual elements to a user are more likely to be trusted~\cite{teach1981analysis}.
We consider explanations as determining \textit{why} a certain decision is consistent with visual evidence, and differentiate between \textit{introspection} explanation systems which explain how a model determines its final output (e.g., ``This is a Western Grebe because filter 2 has a high activation...'') and \textit{justification} explanation systems which produce sentences detailing how visual evidence is compatible with a system output (e.g., ``This is a Western Grebe because it has red eyes...'').
We concentrate on justification explanation systems because such systems may be more useful to non-experts who do not have detailed knowledge of modern computer vision systems~\cite{biran2014justification}.

\begin{figure}[t]
\centering
\includegraphics[height=3.55cm]{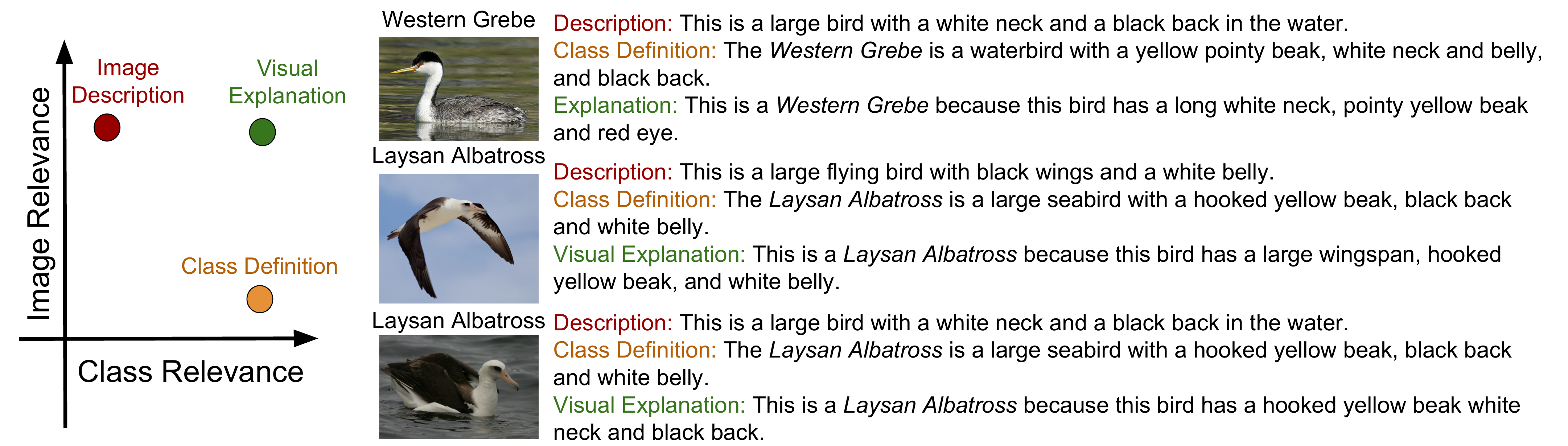}
\vspace{-5pt}
\caption{Our proposed model generates \textit{visual explanations}.  Visual explanations are both image relevant and class relevant.  In contrast, image descriptions are image relevant, but not necessarily class relevant, and class definitions are class relevant but not necessarily image relevant.  In the visual explanations above, class discriminative visual features that are also present in the image are discussed.}
\vspace{-10pt}
\label{fig:teaser}
\end{figure}

We argue that visual explanations must satisfy two criteria: they must both be {\em class discriminative} and {\em  accurately describe} a specific image instance. 
As shown in \autoref{fig:teaser}, explanations are distinct from \textit{descriptions}, which provide a sentence based only on visual information, and \textit{definitions}, which provide a sentence based only on class information. 
Unlike descriptions and definitions, visual explanations detail why a certain category is appropriate for a given image while only mentioning image relevant features.
As an example, let us consider an image classification system that predicts a certain image belongs to the class ``western grebe'' (\autoref{fig:teaser}, top).
A standard captioning system might provide a description such as ``This is a large bird with a white neck and black back in the water.'' %
However, %
as this description does not mention \textit{discriminative} features, %
it could also be applied to a ``laysan albatross'' (\autoref{fig:teaser}, bottom).
In contrast, we propose to provide \textit{explanations}, such as ``This is a western grebe because this bird has a long white neck, pointy yellow beak, and a red eye.'' The explanation includes the ``red eye'' property, \eg, when crucial for distinguishing between ``western grebe'' and ``laysan albatross''. As such, our system explains \textit{why} the predicted category is the most appropriate for the image.

\begin{figure}[t]
\centering

\includegraphics[width=\linewidth]{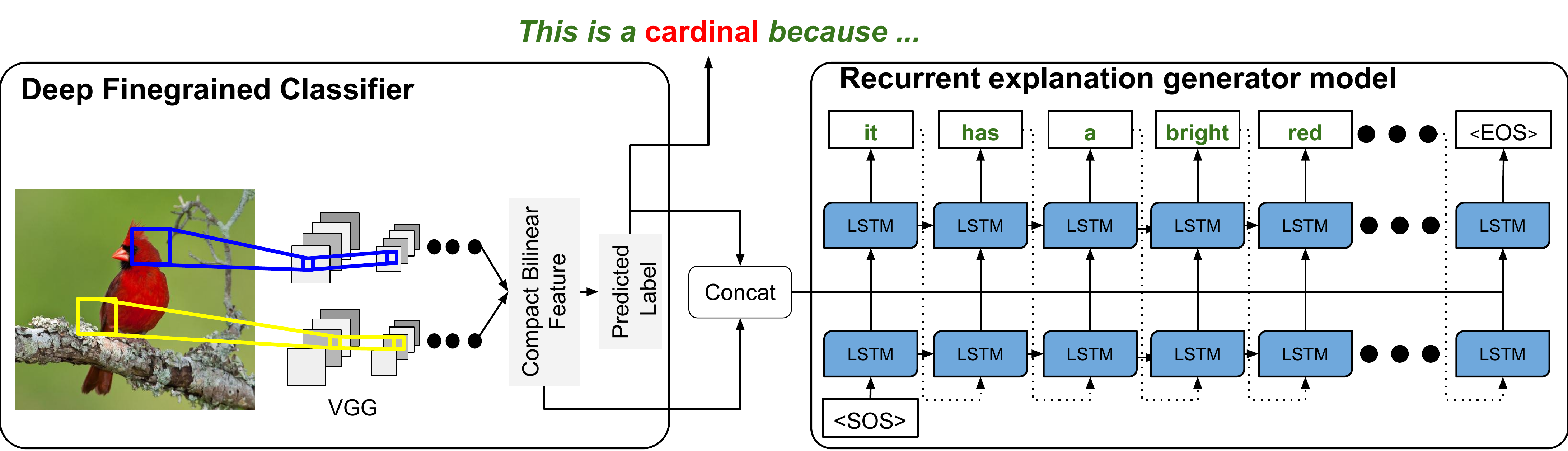}
\vspace{-10pt}
\caption{Generation of explanatory text with our joint classification and language model.  Our model extracts visual features using a fine-grained classifier before language generation.  Additionally, unlike description models we also condition sentence generation on the predicted class label.}
\vspace{-10pt}
\label{fig:teaser-model}
\end{figure}

We outline our approach in Figure~\ref{fig:teaser-model}.
We condition language generation on both an image and a predicted class label which allows us to generate class-specific sentences.
Unlike other caption models,
which condition on visual features from a network pre-trained on ImageNet~\cite{deng2009imagenet},
our model also includes a fine-grained recognition pipeline to produce strong image features~\cite{gao2015compact}.
Like many contemporary description models \cite{VTBE15,donahue15cvpr,KL15,xu2015show,kiros2014multimodal}, our model learns to generate a sequence of words using an LSTM~\cite{HS97}.
However, we design a novel loss function which encourages generated sentences to include class discriminative information.
One challenge in designing a loss to optimize for class specificity is that class specificity is a global sentence property:
\eg, whereas a sentence ``This is an all black bird with a bright red eye'' is class specific to a ``Bronzed Cowbird'', words and phrases in the sentence, such as ``black'' or ``red eye'' are less class discriminative on their own. 
Our proposed generation loss enforces that generated sequences fulfill a certain global property, such as category specificity.
Our final output is a sampled sentence,  so we backpropagate the discriminative loss through the sentence sampling mechanism via a technique from the reinforcement learning literature. 
While typical sentence generation losses optimize the alignment between generated and ground truth sentences, our discriminative loss specifically optimizes for class-specificity. 
To the best of our knowledge, ours is the first method to produce deep visual explanations using natural language justifications. We describe below 
a novel joint vision and language explanation model which combines classification and sentence generation and incorporates a loss function operating over sampled sentences.
We show that this formulation is able to focus generated text to be more discriminative and that our model produces better explanations than a description-only baseline. 
Our results also confirm that generated sentence quality improves with respect to traditional sentence generation metrics by including a discriminative class label loss during training.
This result holds  even when class conditioning is ablated at test time.

\section{Related Work}

\myparagraph{Explanation.} Automatic reasoning and explanation has a long and rich history within the artificial intelligence community \cite{biran2014justification,shortliffe1975model,lane2005explainable,core2006building,van2004explainable,lomas2012explaining,lacave2002review,johnson1994agents}.  Explanation systems span a variety of applications including explaining medical diagnosis \cite{shortliffe1975model}, simulator actions \cite{lane2005explainable,core2006building,van2004explainable,johnson1994agents}, and robot movements \cite{lomas2012explaining}.
Many of these systems are rule-based \cite{shortliffe1975model} or solely reliant on filling in a predetermined template \cite{van2004explainable}.
Methods such as \cite{shortliffe1975model} require expert-level explanations and decision processes.
In contrast, our visual explanation method is learned directly from data by optimizing explanations to fulfill our two proposed visual explanation criteria.
Our model is not provided with expert explanations or decision processes, but rather learns from visual features and text descriptions.  In contrast to systems like \cite{shortliffe1975model,lane2005explainable,core2006building,van2004explainable,lomas2012explaining,lacave2002review} which aim to explain the underlying mechanism behind a decision, authors in \cite{biran2014justification} concentrate on why a prediction is justifiable to a user. 
Such systems are advantageous because they do not rely on user familiarity with the design of an intelligent system in order to provide useful information.

A variety of computer vision methods have focused on discovering visual features which can help ``explain'' an image classification decision \cite{berg2013you,jiang2016learning,doersch2012makes}. 
Importantly, these models do not attempt to link discovered discriminative features to natural language expressions. 
We believe methods to discover discriminative visual features are complementary to our proposed system, as such features could be used as additional inputs to our model and aid  producing better explanations.

\myparagraph{Visual Description.} Early image description methods rely on first detecting visual concepts in a scene (\eg, subject, verb, and object) before generating a sentence with either a simple language model or sentence template \cite{KPD11,guadarrama2013youtube2text}.  
Recent deep models \cite{VTBE15,donahue15cvpr,KL15,xu2015show,kiros2014multimodal,fang2015captions,mao2014explain} have far outperformed such systems and are capable of producing fluent, accurate descriptions of images.
Many of these systems learn to map from images to sentences directly, with no guidance on intermediate features (e.g., prevalent objects in the scene).  
Likewise, our model attempts to learn a visual explanation given only an image and predicted label with no intermediate guidance, such as object attributes or part locations.
Though most description models condition sentence generation only on image features, \cite{jia2015guiding} propose conditioning generation on auxiliary information, such as the words used to describe a similar image in the train set.
However, \cite{jia2015guiding} does not explore conditioning generation on category labels for fine-grained descriptions.

The most common loss function used to train LSTM based sentence generation models \cite{VTBE15,donahue15cvpr,KL15,xu2015show,mao2014explain}
is a cross-entropy loss between the probability distribution of predicted and ground truth words.
Frequently, however, the cross-entropy loss does not directly optimize for  properties that are desired at test time.
\cite{mao16cvpr} proposes an alternative training scheme for generating unambiguous region descriptions which maximizes the probability of a specific region description while minimizing the probability of other region descriptions.
In this work, we propose a novel loss function for sentence generation which allows us to specify a global constraint on generated sentences.

\myparagraph{Fine-grained Classification.}  %
Object classification, and fine-grained classification in particular, is attractive to demonstrate explanation systems because describing image content is not sufficient for an explanation.
Explanation models must focus on aspects that are both class-specific and depicted in the image. 

Most fine-grained zero-shot and few-shot image classification systems use attributes~\cite{LNH13} as auxiliary information that can support visual information. %
Attributes can be thought of as a means to discretize a high dimensional feature space into a series of simple and readily interpretable decision statements that can act as an explanation. However, attributes have several disadvantages. They require fine-grained object experts for annotation which is costly. For each additional class, the list of attributes needs to be revised to ensure discriminativeness so attributes are not generalizable. Finally, though a list of image attributes could help explain a fine-grained classification, attributes do not provide a natural language explanation like the user expects. We therefore, use natural language descriptions collected in \cite{RALS16} which achieved superior performance on zero-shot learning compared to attributes.

\myparagraph{Reinforcement Learning in Computer Vision.}
Vision models which incorporate algorithms from reinforcement learning, specifically how to backpropagate through a sampling mechanism, have recently been applied to visual question answering \cite{andreas16naacl} and activity detection~\cite{yeung2015every}.
Additionally, ~\cite{xu2015show} use a sampling mechanism to attend to specific image regions for caption generation, but use the standard cross-entropy loss during training.

\section{Visual Explanation Model}

\begin{figure}[t]
\centering
\includegraphics[width=\linewidth]{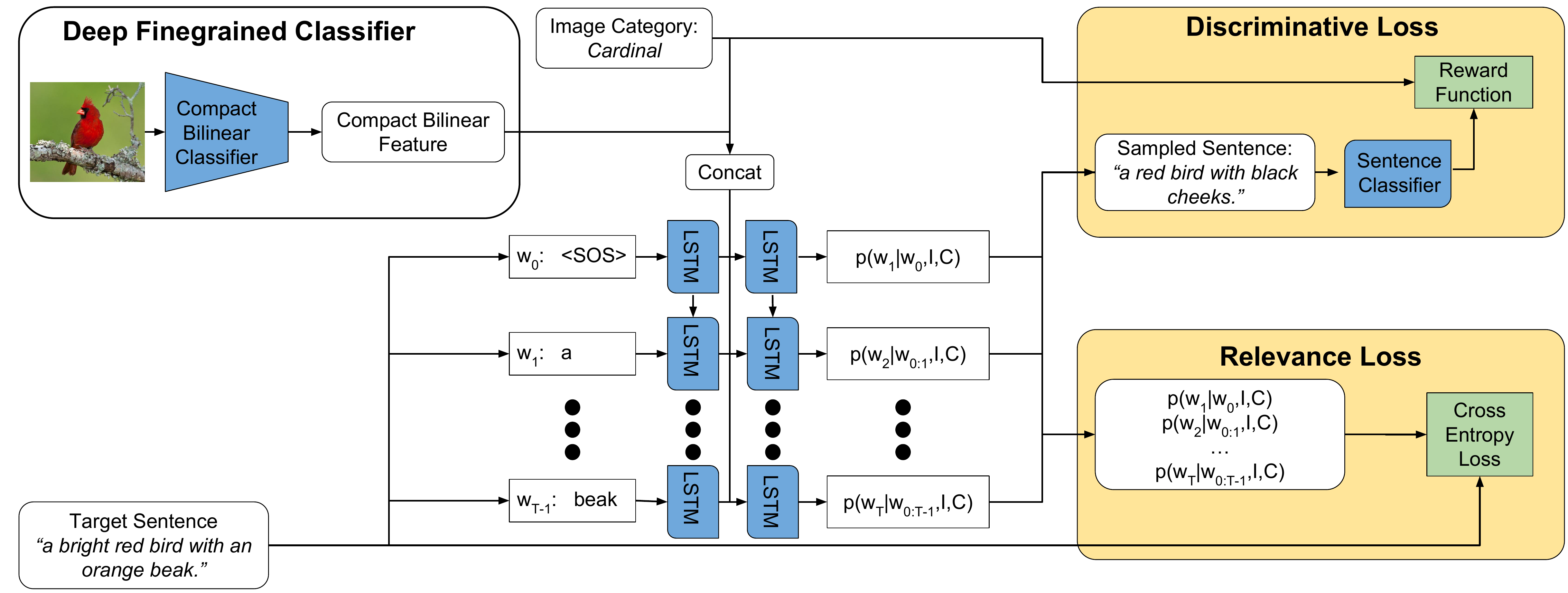}
\caption{Training our explanation model.  Our explanation model differs from other caption models because it (1) includes the object category as an additional input and (2) incorporates a reinforcement learning based discriminative loss}
\vspace{-15pt}
\label{fig:model}
\end{figure}

Our visual explanation model (Figure~\ref{fig:model}) aims to produce an explanation which (1) describes visual content present in a specific image instance and (2) contains appropriate information to explain why an image instance belongs to a specific category.
We ensure generated descriptions meet these two requirements for explanation by including both a \textit{relevance loss} (Figure~\ref{fig:model}, bottom right) and \textit{discriminative loss} (Figure~\ref{fig:model}, top right).
Our main technical contribution is the inclusion of a loss which acts on sampled word sequences during training.
Our proposed loss enables us to enforce global sentence constraints on sentences  and by applying our loss to sampled sentences, we ensure that the final output of our system fulfills our criteria for an explanation. In the following sections we consider a sentence to be a word sequence comprising either a complete sentence or a sentence fragment.

\subsection{Relevance Loss}

Image relevance can be accomplished by training a visual description model.
Our model is based on LRCN~\cite{donahue15cvpr}, which  consists of a convolutional neural network, which extracts powerful high level visual features, and two stacked recurrent networks (specifically LSTMs), which learn how to generate a description conditioned on visual features.
During inference, the first LSTM receives the previously generated word $w_{t-1}$ as input (at time $t=0$ the model receives a ``start-of-sentence'' token), and produces an output $l_{t}$.
The second LSTM, receives the output of the first LSTM $l_{t}$ as well as an image feature $f$ and produces a probability distribution $p(w_t)$ over the next word.  
At each time step, the word $w_t$ is generated by sampling from the distribution $p(w_t)$.
Generation continues until an ``end-of-sentence'' token is generated.

We propose two modifications to the LRCN framework to increase the image relevance of generated sequences (Figure~\ref{fig:model}, top left). 
First, our explanation model uses category predictions as an additional input to the second LSTM in the sentence generation model.
Intuitively, category information can help inform the caption generation model which words and attributes are more likely to occur in a description.
For example, if the caption generation model conditioned only on images mistakes a red eye for a red eyebrow, category level information could indicate the red eye is more likely for a given class. %
We experimented with a few methods to represent class labels, but found a vector representation in which we first train a language model, \eg, an LSTM, to generate word sequences conditioned on images, then compute the average hidden state of the LSTM across all sequences for all classes in the train set worked best.
Second, we use rich category specific features \cite{gao2015compact} to generate relevant explanations. 

Each training instance consists of an image, category label, and a ground truth sentence.  
During training, the model receives the ground truth word $w_t$ for each time step $t \in T$.
We define the relevance loss as:
\begin{align*}
L_R = \frac{1}{N} \sum_{n=0}^{N-1}\sum_{t=0}^{T-1} \log p(w_{t+1}|w_{0:t}, I, C)
\end{align*}
where $w_t$ is a ground truth word, $I$ is the image, $C$ is the category, and $N$ is the batch size.
By training the model to predict each word in a ground truth sentence, the model is trained to produce sentences which correspond to image content.
However, this loss does not explicitly encourage generated sentences to discuss discerning visual properties.
In order to generate sentences which are both image relevant and category specific, we include a discriminative loss to focus sentence generation on discriminative visual properties of an image.

\subsection{Discriminative Loss}

Our discriminative loss is based on %
a reinforcement learning paradigm for learning with layers which require intermediate activations of a network to be sampled.
In our formulation, we first sample a sentence and then input the sampled sentence into a discriminative loss function.
By sampling the sentence before computing the loss,
we ensure that sentences sampled from our model are more likely to be class discriminative.
We first overview how to backpropagate through the sampling mechanism, then discuss how we calculate the discriminative loss.

The overall function we minimize in the explanation network weights
$W$ is
$
L_R -
\lambda
\mathbb{E}_{\tilde{w} \sim p(w)}
\left[
R_D(\tilde{w})
\right]
$,
a linear combination of the relevance loss $L_R$ and the expectation of the negative discriminator reward $-R_D(\tilde{w})$ over descriptions $\tilde{w} \sim p(w|I,C)$, where $p(w|I,C)$ is the model's estimated conditional distribution over descriptions $w$ given the image $I$ and category $C$.
Since this expectation over descriptions is intractable, we estimate it at training time using Monte Carlo sampling of descriptions from the categorical distribution given by the model's softmax output at each timestep.
As a discrete distribution, the sampling operation for the categorical distribution is non-smooth in the distribution's parameters $\{p_i\}$, so the gradient $
\nabla_W R_D(\tilde{w})
$
of the reward $R_D$ for a given sample $\tilde{w}$ with respect to the weights $W$ is undefined.

Following REINFORCE~\cite{reinforce}, we make use of the following equivalence property of the expected reward gradient:
\begin{equation*}
\nabla_W
\mathbb{E}_{\tilde{w} \sim p(w)}
\left[
R_D(\tilde{w})
\right]
=
\mathbb{E}_{\tilde{w} \sim p(w)}
\left[
R_D(\tilde{w})
\nabla_W
\log p(\tilde{w})
\right]
\end{equation*}
In the reformulation on the right-hand side, the gradient $
\nabla_W \log p(\tilde{w})
$ is well-defined:
$\log p(\tilde{w})$ is the log-likelihood of the sampled description $\tilde{w}$,
just as $L_R$ was the log-likelihood of the ground truth description.
In this case, however, the sampled gradient term is weighted by the reward $R_D(\tilde{w})$, pushing the weights to increase the likelihood assigned to the most highly rewarded (and hence most discriminative) descriptions.

Therefore, the final gradient we compute to update the weights $W$, given a description $\tilde{w}$ sampled from the model's softmax distribution, is:
\begin{equation*}
\nabla_W
L_R
-
\lambda
R_D(\tilde{w})
\nabla_W
\log p(\tilde{w}) .
\end{equation*}
$R_D(\tilde{w})$ should be high when sampled sentences are discriminative.
We define our reward simply as $R_D(\tilde{w}) = p(C|\tilde{w})$, or the probability of the ground truth category $C$ given only the generated sentence $\tilde{w}$.
By placing the discriminative loss after the sampled sentence, the sentence acts as an information bottleneck.
For the model to produce an output with a large reward, the generated sentence must include enough information to classify the original image properly.
For the sentence classifier, we train a single layer LSTM-based classification network to classify ground truth sentences.
Our sentence classifier correctly predicts the class of unseen validation set sentences $22\%$ of the time.
This number is possibly low because descriptions in the dataset do not necessarily contain discriminative properties (e.g.,  ``This is a white bird with grey wings.'' is a valid description but can apply to multiple bird species).
Nonetheless, we find that this classifier provides enough information to train our explanation model.
We do not update the sentence classifier weights when training our explanation model.

\section{Experimental Setup}
\label{sec:exp_setup}

\myparagraph{Dataset.} In this work, we employ the Caltech UCSD Birds 200-2011 (CUB) dataset~\cite{WahCUB_200_2011} which contains 200 classes of North American bird species and 11,788 images in total. 
A recent extension to this dataset~\cite{RALS16} collected 5 sentences for each of the images. These sentences do not only describe the content of the image, e.g., ``This is a bird'', but also gives a detailed description of the bird, e.g., ``that has a cone-shaped beak, red feathers and has a black face patch''.
Unlike other image-sentence datasets, every image in the CUB dataset belongs to a class, and therefore sentences as well as images are associated with a single label. 
This property makes this dataset unique for the visual explanation task, where our aim is to generate sentences that are both discriminative and class-specific. 
We stress that sentences collected in~\cite{RALS16} were not collected for the task of visual explanation.
Consequently, they do not explain why an image belongs to a certain class, but rather include discriptive details about each bird class.

\myparagraph{Implementation.} For image features, we extract 8,192 dimensional features from the penultimate layer of the compact bilinear fine-grained classification model \cite{gao2015compact} which has been pre-trained on the CUB dataset and achieves an accuracy of $84\%$.
We use one-hot vectors to represent input words at each time step and learn a $1,000$-dimensional embedding before inputting each word into the a 1000-dimensional LSTM.
We train our models using \textit{Caffe}~\cite{jia2014caffe}, and determine model hyperparameters using the standard CUB validation set before evaluating on the test set.  All reported results are on the standard CUB test set.  %

\myparagraph{Baseline and Ablation Models.}
In order to investigate our explanation model, we propose two baseline models: a \textit{description} model and a \textit{definition} model.
Our description baseline is trained to generate sentences conditioned only on images and is equivalent to LRCN \cite{donahue15cvpr} except we use features from a fine-grained classifier.
Our definition model is trained to generate sentences using only the image label as input.  
Consequently, this model outputs the same sentence for different image instances of the same class.
By comparing these baselines to our explanation model, we demonstrate that our explanation model is both more image and class relevant, and thus generates superior explanations.

Our explanation model differs from a description model in two key ways.  
First, in addition to an image, generated sentences are conditioned on class predictions.
Second, our explanations are trained with a discriminative loss which enforces that generated sentences contain class specific information.
To understand the importance of these two contributions, we compare our explanation model to an \textit{explanation-label} model which is not trained with the discriminative loss, and to an \textit{explanation-discriminative} model which is not conditioned on the predicted class.
By comparing our explanation model to the explanation-label model and explanation-discriminative model, we demonstrate that both class information and the discriminative loss are important in generating descriptions.

\myparagraph{Metrics.} To evaluate our explanation model, we use both automatic metrics and a human evaluation. 
Our automatic metrics rely on the common sentence evaluation metrics, METEOR~\cite{banerjee2005meteor} and CIDEr~\cite{vedantam2015cider}. 
METEOR is computed by matching words in generated and reference sentences, but unlike other common metrics such as BLEU~\cite{papineni2002bleu}, uses WordNet~\cite{miller1990introduction} to also match synonyms.  
CIDEr measures the similarity of a generated sentence to reference sentence by counting common n-grams which are TF-IDF weighted.
Consequently, the metric rewards sentences for correctly including n-grams which are uncommon in the dataset.

A generated sentence is \textit{image relevant} if it mentions concepts which are mentioned in ground truth reference sentences for the image.
Thus, to measure image relevance we simply report METEOR and CIDEr scores, with more relevant sentences producing higher METEOR and CIDEr scores.

Measuring \textit{class relevance} is considerably more difficult.
We could use the LSTM sentence classifier used to train our discriminative loss, but this is an unfair metric because some models were trained to directly increase the accuracy as measured by the LSTM classifier.
Instead, we measure class relevance by considering how similar generated sentences for a class are to ground truth sentences for that class.
Sentences which describe a certain bird class, e.g., ``cardinal'', should contain similar words and phrases to ground truth ``cardinal'' sentences, but not ground truth ``black bird'' sentences.
We compute CIDEr scores for images from each bird class, but instead of using ground truth image descriptions as reference sentences, we use all reference sentences which correspond to a particular class.
We call this metric the \textit{class similarity} metric.

More class relevant sentences should result in a higher CIDEr scores, but it is possible that if a model produces better overall sentences it will have a higher CIDEr score without generating more class relevant descriptions.
To further demonstrate that our sentences are class relevant, we also compute a \textit{class rank} metric.
To compute this metric, we compute the CIDEr score for each generated sentence and use ground truth reference sentences from each of the 200 classes in the CUB dataset as references.
Consequently, each image is associated with a CIDEr score which measures the similarity of the generated sentences to each of the 200 classes in the CUB dataset.
CIDEr scores computed for generated sentences about cardinals should be higher when compared to cardinal reference sentences than when compared to reference sentences from other classes.

We choose to emphasize the CIDEr score when measuring class relevance because it includes the TF-IDF weighting over n-grams.
Consequently, if a bird includes a unique feature, such as ``red eyes'', generated sentences which mention this attribute should be rewarded more than sentences which just mention attributes common across all bird classes.

The ultimate goal of an explanation system is to provide useful information to a human.
We therefore also consulted %
experienced bird watchers to rate our explanations against our two baseline and ablation models.
We provided a random sample of images in our test set with sentences generated from each of our five models and asked the bird watchers to rank which sentence explained the classification best.
Consulting experienced bird watchers is important because some sentences may list correct, but non-discriminative, attributes.
For example, a sentence ``This is a Geococcyx because this bird has brown feathers and a brown crown.'' may be a correct description, but if it does not mention unique attributes of a bird class, it is a poor explanation.
Though it is difficult to expect an average person to infer or know this information, experienced bird watchers are aware of which features are important in bird classification.

\section{Results}
\label{sec:results}

We demonstrate that our model produces visual explanations by showing that our generated explanations fulfill the two aspects of our proposed definition of visual explanation and are image relevant and class relevant.
Furthermore, we demonstrate that by training our model to generate class specific descriptions, we generate higher quality sentences based on common sentence generation metrics.

\begin{table}[t]
 \begin{center}
 \caption{Comparison of our explanation model to our definition and description baseline, as well as the explanation-label and explanation-discriminative (explanation-dis. in the table) ablation models.  We demonstrate that our generated explanations are image relevant by computing METEOR and CIDEr scores (higher is better).  We demonstrate class relevance using a class similarity metric (higher is better) and class rank metric (lower is better) (see Section~\ref{sec:exp_setup} for details).  Finally, we ask experienced bird watchers to rank our explanations.  On all metrics, our explanation model performs best.}
  \begin{tabular}{|l|c c |c c | c |}
  \hline
    & \multicolumn{2}{c|}{\textbf{Image Relevance}} & \multicolumn{2}{c|}{\textbf{Class Relevance}} & \textbf{Best Explanation}\\ 
	& METEOR & CIDEr & Similarity & Rank & Bird Expert Rank\\
    & & & & (1-200)& (1-5)\\
	\hline
	Definition & $27.9$ & $43.8$ & $42.60$ &  $15.82$ & 2.92 \\%2.83\\
    Description & $27.7$ & $42.0$ & $35.3$ & $24.43$ & 3.11 \\%3.13\\
    \hline
    Explanation-Label & $28.1$ & $44.7$ & $40.86$ & $17.69$ & 2.97\\%3.13\\
    Explanation-Dis. & $28.8$ & $51.9$ & $43.61$ & $19.80$ & 3.22\\%3.30\\
    Explanation& \textbf{29.2} & \textbf{56.7}  & \textbf{52.25} &  \textbf{13.12} &  \textbf{2.78} \\%2.48}\\
    \hline
  \end{tabular}
 \end{center}
\vspace{-20pt}
\label{tab:results}
\end{table}

\subsection{Quantitative Results}

\myparagraph{Image Relevance.}
Table~\ref{tab:results}, columns 2 \& 3, record METEOR and CIDEr scores for our generated sentences.
Importantly, our explanation model has higher METEOR and CIDEr scores than our baselines.  The explanation model also outperforms the explanation-label and explanation-discriminative model suggesting that both label conditioning and the discriminative loss are key to producing better sentences.
Furthermore, METEOR and CIDEr are substantially higher when including a discriminative loss during training (compare rows 2 and 4 and rows 3 and 5) demonstrating that including this additional loss leads to better generated sentences.
Surprisingly, the definition model produces more image relevant sentences than the description model. %
Information in the label vector and image appear complimentary as the explanation-label model, which conditions generation both on the image and label vector, produces better sentences.

\myparagraph{Class Relevance.}
Table~\ref{tab:results}, columns 4 \& 5, record the class similarity and class rank metrics (see Section~\ref{sec:exp_setup} for details).
Our explanation model produces a higher class similarity score than other models by a substantial margin.
The class rank for our explanation model is  also lower than for any other model suggesting that sentences generated by our explanation model more closely resemble the correct class than other classes in the dataset.
We emphasize that our goal is to produce reasonable explanations for classifications, not rank categories based on our explanations.
We expect the rank of sentences produced by our explanation model to be lower, but not necessarily rank one.
Our ranking metric is quite difficult; sentences must include enough information to differentiate between very similar bird classes without looking at an image, and our results clearly show that our explanation model performs best at this difficult task.
Accuracy scores produced by our LSTM sentence classifier follow the same general trend, with our explanation model producing the highest accuracy ($59.13\%$) and the description model producing the lowest accuracy ($22.32\%$).

\myparagraph{Explanation.} Table~\ref{tab:results}, column 6 details the evaluation of two experienced bird watchers.  
The bird experts evaluated 91 randomly selected images and answered which sentence provided the best explanation for the bird class.
Our explanation model has the best mean rank (lower is better), followed by the description  model.
This trend resembles the trend seen when evaluating class relevance.
Additionally, all models which are conditioned on a label (lines 1, 3, and 5) have lower rank suggesting that label information is important for explanations.

\subsection{Qualitative Results}

\begin{figure}[t]
\centering
\includegraphics[width=\linewidth]{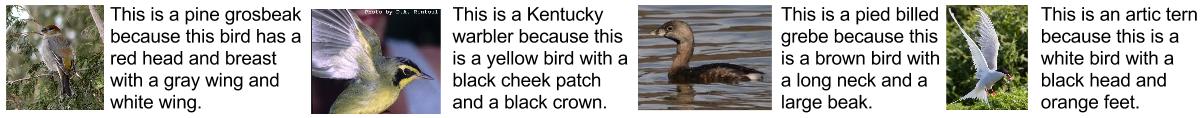}
\vspace{-20pt}
\caption{Visual explanations generated by our system.  Our explanation model produces image relevant sentences that also discuss class discriminative attributes.}
\label{fig:exampleExplanations}
\vspace{-15pt}
\end{figure}

Figure~\ref{fig:exampleExplanations} shows sample explanations produced by first outputing a declaration of the predicted class label (``This is a warbler...") and then a justification conjunction (e.g., ``because") followed by the explantory text sentence fragment produced by the model described above in Section 3.
Qualitatively, our explanation model performs quite well. 
Note that our model accurately describes fine detail such as ``black cheek patch'' for ``Kentucky warbler'' and ``long neck'' for ``pied billed grebe''. %
For the remainder of our qualitative results, we omit the class declaration for easier comparison.

\begin{figure}[h]
\centering
\includegraphics[width=\linewidth]{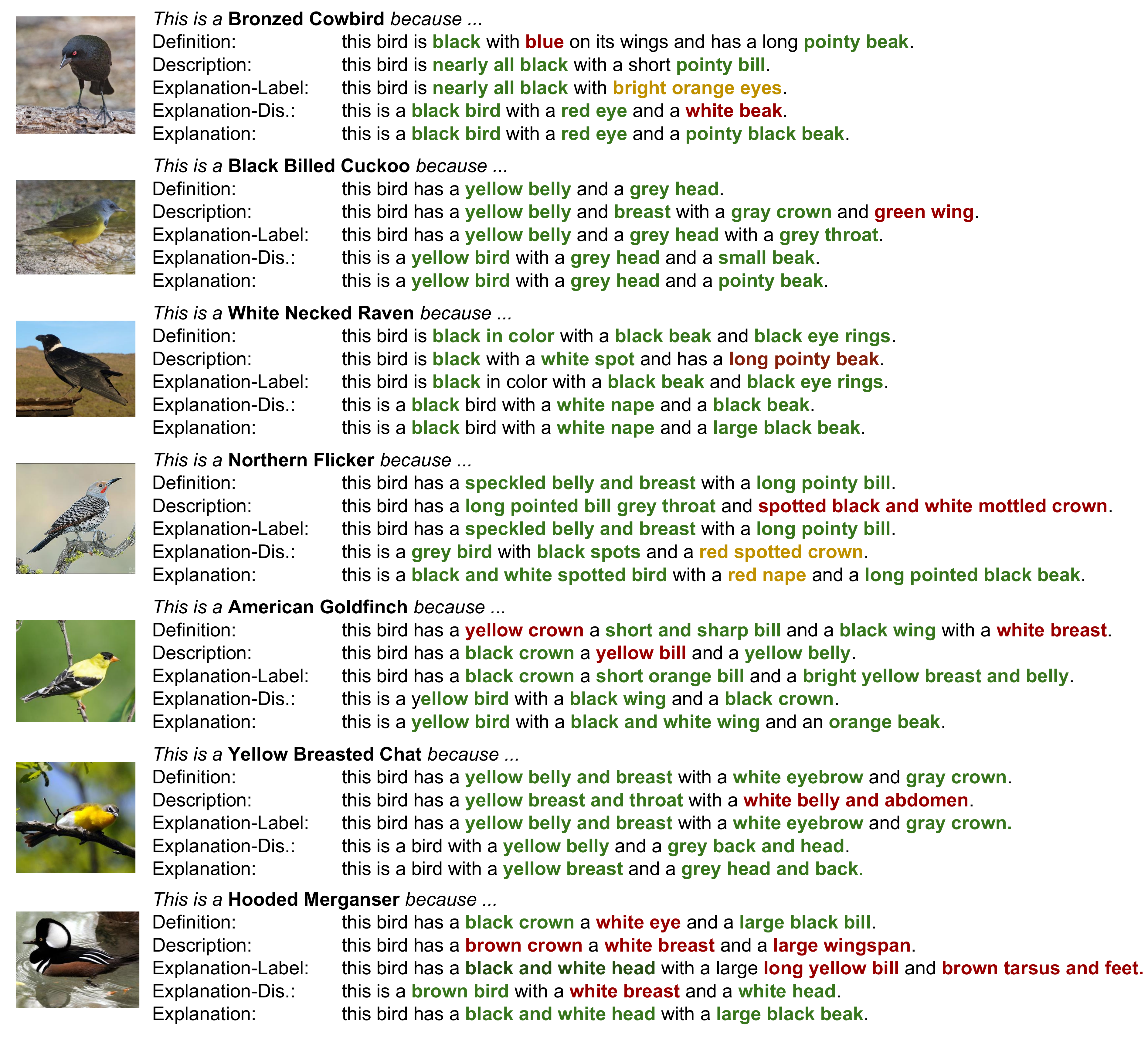}
\vspace{-10pt}
\caption{Example sentences generated by our baseline models, ablation models, and proposed explanation model.  Correct attributes are highlighted in green, mostly correct attributes are highlighted in yellow, and incorrect attributes are highlighted in red.  The explanation model consistently discusses image relevant and class relevant features.}
\label{fig:compare5}
\vspace{-10pt}
\end{figure}

\myparagraph{Comparison of Explanations, Baselines, and Ablations.}
Figure~\ref{fig:compare5} compares sentences generated by our definition and description baselines, explanation-label and explanation-discriminative ablations and explanation model.
Each model produces reasonable sentences, however, we expect our explanation model to produce sentences which discuss class relevant attributes.
For many images, the explanation model mentions attributes that not all other models mention.
For example, in Figure~\ref{fig:compare5}, row 1, the explanation model specifies that the ``bronzed cowbird'' has ``red eyes'' which is a rarer bird attribute than attributes mentioned correctly by the definition and description models (``black'', ``pointy bill'').
Similarly, when explaining the ``White Necked Raven'' (Figure~\ref{fig:compare5} row 3), the explanation model identifies the ``white nape'', which is a unique attribute of that bird.
Based on our image relevance metrics, we also expect our explanations to be more image relevant.  
An obvious example of this is in Figure~\ref{fig:compare5} row 7 where the explanation model includes only attributes present in the image of the ``hooded merganser'', whereas all other models mention at least one incorrect attribute.

\begin{figure}[t]
\centering
\includegraphics[width=\linewidth]{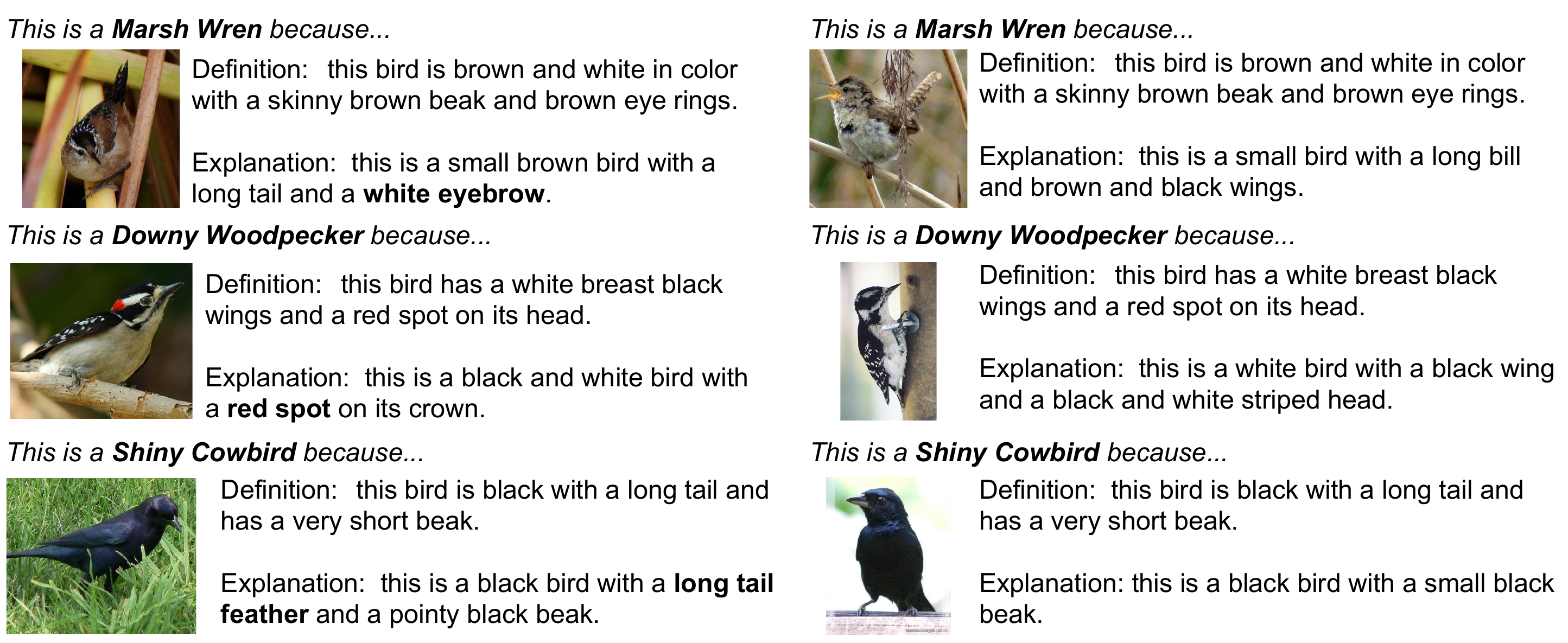}
\vspace{-5pt}
\caption{We compare generated explanations and descriptions.  All explanations on the left include an attribute which is not present on the image on the right.  In contrast to definitions, our explanation model can adjust its output based on visual evidence.
}
\label{fig:explanationVDescription}
\vspace{-10pt}
\end{figure}

\myparagraph{Comparing Definitions and Explanations.} Figure~\ref{fig:explanationVDescription} directly compares explanations to definitions for three bird categories.  
Explanations in the left column include an attribute about an image instance of a bird class which is not present in the image instance of the same bird class in the right column.
Because the definition remains constant for all image instances of a bird class, the definition can produce sentences which are not image relevant.
For example, in the second row, the definition model indicates that the bird has a ``red spot on its head''.  
Though this is true for the image on the left and for many ``Downy Woodpecker'' images, it is not true for the image on the right.
In contrast, the explanation model produces image relevant sentences for both images.

\begin{figure}[t]
\centering
\includegraphics[width=\linewidth]{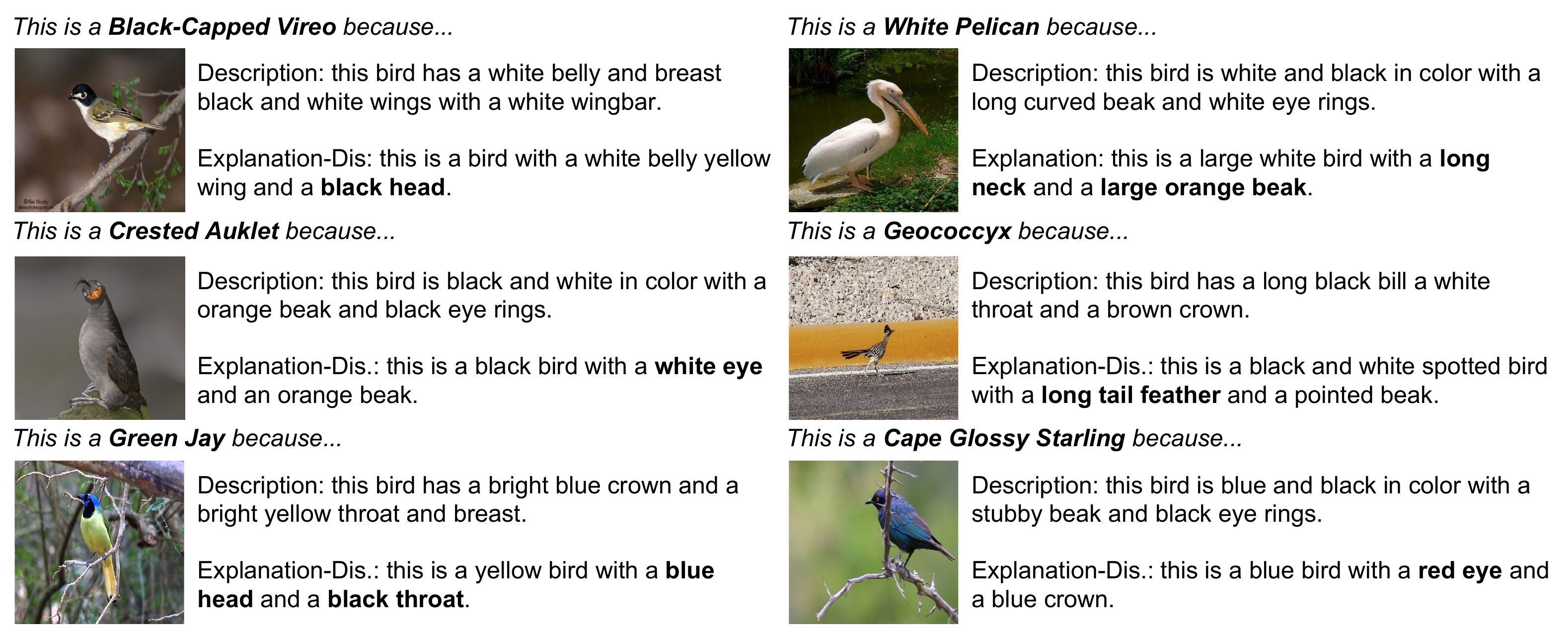}
\vspace{-5pt}
\caption{Comparison of sentences generated using description and explanation-discriminative models. Though both are capable of accurately describing visual attributes, the explanation-discriminative model captures more ``class-specific'' attributes.}
\vspace{-5pt}
\label{fig:discrim}
\end{figure}

\myparagraph{Training with the Discriminative Loss.} To illustrate how the discriminative loss impacts sentence generation we directly compare the description model to the explanation-discriminative model in Figure~\ref{fig:discrim}.  
Neither of these models receives class information at test time, though the explanation-discriminative model is explicitly trained to produced class specific sentences.
Both models can generate visually correct sentences. 
However, generated sentences trained with our discriminative loss contain properties specific to a class more often than the ones generated using the image description model, even though neither has access to the class label at test time. 
For instance, for the class ``black-capped vireo'' both models discuss properties which are visually correct, but the explanation-discriminative model mentions ``black head'' which is one of the most prominent distinguishing properties of this vireo type. 
Similarly, for the ``white pelican'' image, the explanation-discriminative model mentions the properties ``long neck'' and ``orange beak'', which are fine-grained and discriminative. 

\begin{figure}[t]
\centering
\includegraphics[width=\linewidth]{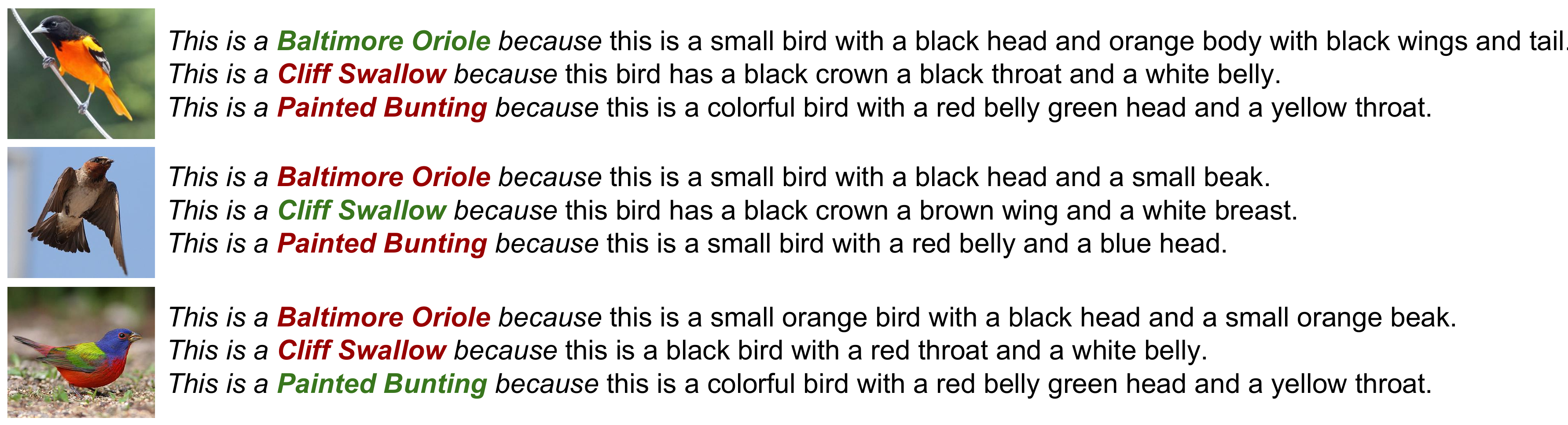}
\vspace{-10pt}
\caption{We observe how explanations change when conditioning on different classes.  Some bird categories, like ``painted bunting'' carry strong class information that heavily influence the explanation.}
\vspace{-10pt}
\label{fig:CapCond}
\end{figure}

\myparagraph{Class Conditioning.} To qualitatively observe the relative importance of image features and label features in our explanation model, we condition explanations for a ``baltimore oriole'', ``cliff swallow'', and ``painted bunting'' on the correct class and incorrect classes (Figure~\ref{fig:CapCond}).
When conditioning on the ``painted bunting'', the explanations for ``cliff swallow'' and ``baltimore oriole'' both include colors which are not present suggesting that the ``painted bunting'' label encourages generated captions to include certain color words.
However, for the ``baltimore oriole'' image, the colors mentioned when conditioning on ``painted bunting'' (red and yellow) are similar to the true color of the oriole (yellow-orange) suggesting that visual evidence informs sentence generation.

\section{Conclusion}

Explanation is an important capability for deployment of intelligent systems.
Visual explanation is a rich research direction, especially as the field of computer vision continues to employ and improve deep models which are not easily interpretable.
Our work is an important step towards explaining deep visual models.
We anticipate that future models will look ``deeper'' into networks to produce explanations and perhaps begin to explain the internal mechanism of deep models.

To build our explanation model, we proposed a novel reinforcement learning based loss which allows us to influence the kinds of sentences generated with a sentence level loss function.
Though we focus on a discriminative loss in this work, we believe the general principle of including a loss which operates on a sampled sentence and optimizes for a global sentence property is potentially beneficial in other applications.  
For example, \cite{hendricks16cvpr,mao2015learning} propose introducing new vocabulary words into a captioning system.
Though both models aim to optimize a global sentence property (whether or not a caption mentions a certain concept), neither optimizes for this property directly.

In summary, we have presented a novel framework which provides explanations of a visual classifier.  
Our quantitative and qualitative evaluations demonstrate the potential of our proposed model and effectiveness of our novel loss function.
Our explanation model goes beyond the capabilities of current captioning systems and effectively incorporates classification information to produce convincing explanations, a potentially key advance for adoption of many sophisticated AI systems.

\paragraph{Acknowledgements.}
 This work was supported by DARPA, AFRL, DoD MURI award N000141110688, NSF awards IIS-1427425 and IIS-1212798, and the Berkeley Vision and Learning Center.  Marcus Rohrbach was supported by a fellowship within the FITweltweit-Program of the German Academic Exchange Service (DAAD).
 Lisa Anne Hendricks is supported by an NDSEG fellowship.
We thank our experienced bird watchers, Celeste Riepe and Samantha Masaki, for helping us evaluate our model. 

\bibliographystyle{splncs}
\bibliography{biblioLong,egbib}
\end{document}

%% file: defines.tex
 \makeatletter
 \DeclareRobustCommand\onedot{\futurelet\@let@token\@onedot}
 \def\@onedot{\ifx\@let@token.\else.\null\fi\xspace}
 \def\eg{e.g\onedot} \def\Eg{E.g\onedot}
 \def\ie{i.e\onedot} \def\Ie{I.e\onedot}
 \def\cf{cf\onedot} \def\Cf{Cf\onedot}
 \def\etc{etc\onedot} \def\vs{vs\onedot}
 \def\wrt{w.r.t\onedot} \def\dof{d.o.f\onedot}
 \def\etal{\textit{et~al\onedot}} \def\iid{i.i.d\onedot}
 \def\Fig{Fig\onedot} \def\Eqn{Eqn\onedot} \def\Sec{Sec\onedot}
 \def\vs{vs\onedot}
 \makeatother

\DeclareRobustCommand{\figref}[1]{Fig.~\ref{#1}}
\DeclareRobustCommand{\figsref}[1]{Figures~\ref{#1}}

\DeclareRobustCommand{\Figref}[1]{Fig.~\ref{#1}}
\DeclareRobustCommand{\Figsref}[1]{Figures~\ref{#1}}

\DeclareRobustCommand{\Secref}[1]{Section~\ref{#1}}
\DeclareRobustCommand{\secref}[1]{Section~\ref{#1}}

\DeclareRobustCommand{\Secsref}[1]{Sections~\ref{#1}}
\DeclareRobustCommand{\secsref}[1]{Sections~\ref{#1}}

\DeclareRobustCommand{\Tableref}[1]{Table~\ref{#1}}
\DeclareRobustCommand{\tableref}[1]{Table~\ref{#1}}

\DeclareRobustCommand{\Tablesref}[1]{Tables~\ref{#1}}
\DeclareRobustCommand{\tablesref}[1]{Tables~\ref{#1}}

\DeclareRobustCommand{\eqnref}[1]{Equation~(\ref{#1})}
\DeclareRobustCommand{\Eqnref}[1]{Equation~(\ref{#1})}

\DeclareRobustCommand{\eqnsref}[1]{Equations~(\ref{#1})}
\DeclareRobustCommand{\Eqnsref}[1]{Equations~(\ref{#1})}

\DeclareRobustCommand{\chapref}[1]{Chapter~\ref{#1}}
\DeclareRobustCommand{\Chapref}[1]{Chapter~\ref{#1}}

\DeclareRobustCommand{\chapsref}[1]{Chapters~\ref{#1}}
\DeclareRobustCommand{\Chapsref}[1]{Chapters~\ref{#1}}

\hyphenation{po-si-tive}

%% file: macros.tex
\newcommand{\scream}[1]{\textbf{*** #1! ***}}
 \newcommand{\fixme}[1]{\textcolor{red}{\textbf{FiXme}#1}\xspace}
 \newcommand{\hobs}{\textrm{h}_\textrm{obs}}
 \newcommand{\cpad}[1]{@{\hspace{#1mm}}}
 \newcommand{\alg}[1]{\textsc{#1}}

 \newcommand{\fnrot}[2]{\scriptsize\rotatebox{90}{\begin{minipage}{#1}\flushleft #2\end{minipage}}}
 \newcommand{\chmrk}{{\centering\ding{51}}}
 \newcommand{\eqn}[1]{\begin{eqnarray}\vspace{-1mm}#1\vspace{-1mm}\end{eqnarray}}
 \newcommand{\eqns}[1]{\begin{eqnarray*}\vspace{-1mm}#1\vspace{-1mm}\end{eqnarray*}}

\newcommand{\todo}[1]{\textcolor{red}{ToDo: #1}}
\newcommand{\myparagraph}[1]{\noindent \textbf{#1}}

\newcommand{\invisible}[1]{}%

\newcommand{\figvspace}{\vspace{-.5cm}}
\newcommand{\secvspace}{\vspace{-.2cm}}
\newcommand{\subsecvspace}{\vspace{-.2cm}}

\newcommand{\approach}{GroundeR\xspace}

\graphicspath{{./fig/}{./fig/plots/}}